\documentclass[letterpaper]{article} % DO NOT CHANGE THIS
\usepackage[draft]{aaai2026}  % DO NOT CHANGE THIS
\usepackage{times}  % DO NOT CHANGE THIS
\usepackage{helvet}  % DO NOT CHANGE THIS
\usepackage{courier}  % DO NOT CHANGE THIS
\usepackage[hyphens]{url}  % DO NOT CHANGE THIS
\usepackage{graphicx} % DO NOT CHANGE THIS
\usepackage{natbib}  % DO NOT CHANGE THIS AND DO NOT ADD ANY OPTIONS TO IT
\usepackage{caption} % DO NOT CHANGE THIS AND DO NOT ADD ANY OPTIONS TO IT
\usepackage{algorithm}
\usepackage{algorithmic}
\usepackage{multirow}
\usepackage{booktabs}
\usepackage{amssymb}
\usepackage{bbding}
\usepackage[table]{xcolor}
\definecolor{first}{RGB}{191, 225, 201} % 第一种绿色
\definecolor{second}{RGB}{227, 237, 185} % 第二种黄色绿色
\definecolor{third}{RGB}{254, 250, 194} % 第三种浅黄色
\usepackage{amsmath}

\usepackage{newfloat}
\usepackage{listings}
\DeclareCaptionStyle{ruled}{labelfont=normalfont,labelsep=colon,strut=off} % DO NOT CHANGE THIS
\floatstyle{ruled}
\newfloat{listing}{tb}{lst}{}
\floatname{listing}{Listing}
\title{WDT-MD: Wavelet Diffusion Transformers for Microaneurysm \\Detection in Fundus Images}
\author{
    Yifei Sun\textsuperscript{\rm 1, 2}, Yuzhi He\textsuperscript{\rm 3}, Junhao Jia\textsuperscript{\rm 2}, Jinhong Wang\textsuperscript{\rm 1}, \\
    Ruiquan Ge\textsuperscript{\rm 2}\thanks{Corresponding authors.}, Changmiao Wang\textsuperscript{\rm 4}$^*$, Hongxia Xu\textsuperscript{\rm 1, 5}$^*$
}
\affiliations{
    \textsuperscript{\rm 1}Transvascular Implantation Devices Research Institute, Zhejiang University, Hangzhou, China\\
    \textsuperscript{\rm 2}Hangzhou Dianzi University, Hangzhou, China\\
    \textsuperscript{\rm 3}Xidian University, Xi'an, China\\
    \textsuperscript{\rm 4}Shenzhen Research Institute of Big Data, Shenzhen, China\\
    \textsuperscript{\rm 5}WeDoctor Cloud and Liangzhu Laboratory, Hangzhou, China\\
    \{diaoquesang, cmwangalbert\}@gmail.com, 23012100059@stu.xidian.edu.cn, \\
    \{23080631, gespring\}@hdu.edu.cn, \{wangjinhong, einstein\}@zju.edu.cn
}

\begin{document}

\maketitle

\begin{abstract}
Microaneurysms (MAs), the earliest pathognomonic signs of Diabetic Retinopathy (DR), present as sub-60 $\mu m$ lesions in fundus images with highly variable photometric and morphological characteristics, rendering manual screening not only labor-intensive but inherently error-prone. While diffusion-based anomaly detection has emerged as a promising approach for automated MA screening, its clinical application is hindered by three fundamental limitations. First, these models often fall prey to ``identity mapping", where they inadvertently replicate the input image. Second, they struggle to distinguish MAs from other anomalies, leading to high false positives. Third, their suboptimal reconstruction of normal features hampers overall performance. To address these challenges, we propose a Wavelet Diffusion Transformer framework for MA Detection (WDT-MD), which features three key innovations: a noise-encoded image conditioning mechanism to avoid ``identity mapping" by perturbing image conditions during training; pseudo-normal pattern synthesis via inpainting to introduce pixel-level supervision, enabling discrimination between MAs and other anomalies; and a wavelet diffusion Transformer architecture that combines the global modeling capability of diffusion Transformers with multi-scale wavelet analysis to enhance reconstruction of normal retinal features. Comprehensive experiments on the IDRiD and e-ophtha MA datasets demonstrate that WDT-MD outperforms state-of-the-art methods in both pixel-level and image-level MA detection. This advancement holds significant promise for improving early DR screening.
\end{abstract}

% Uncomment the following to link to your code, datasets, an extended version or similar.
% You must keep this block between (not within) the abstract and the main body of the paper.
\begin{links}
    \link{Code}{https://github.com/diaoquesang/WDT-MD}
    % \link{Datasets}{https://aaai.org/example/datasets}
    % \link{Extended version}{https://aaai.org/example/extended-version}
\end{links}

\section{Introduction}
\begin{figure}[!t]
\centering
\includegraphics[width=0.95\columnwidth]{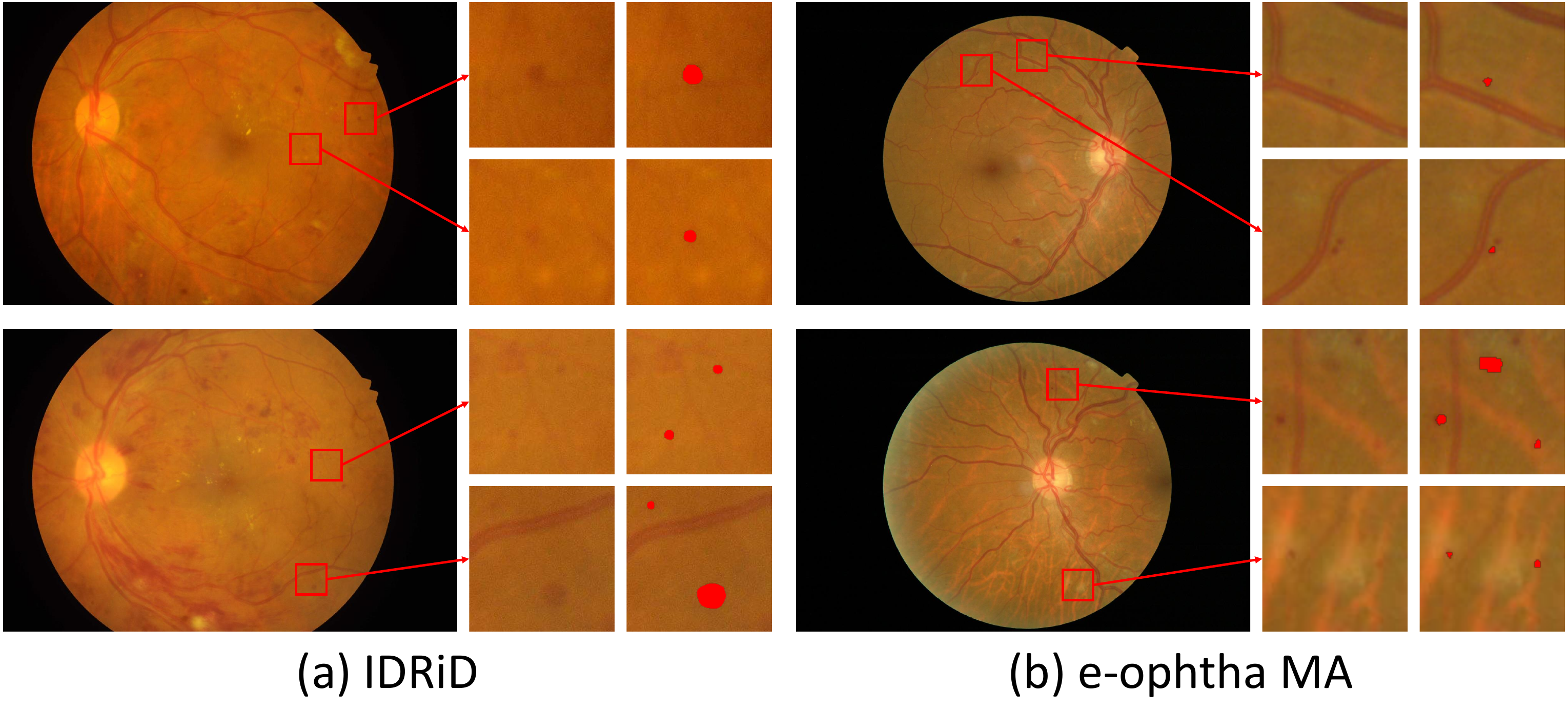}
\caption{An illustration of MAs in fundus images. (a) is sampled from the IDRiD dataset \cite{porwal2018indian}, and (b) is from the e-ophtha MA dataset \cite{decenciere2013teleophta}. Three columns in each sub-figure depict the fundus image, patches zooming in MAs, and MA areas marked in red, respectively. Most MAs are within 60 $\mu m$ in diameter, close to 6 pixels in a fundus image with 10 $\mu m$ pixel spacing.}
\label{data}
\end{figure}

Diabetic Retinopathy (DR) is a serious complication affecting individuals with diabetes and can result in severe vision loss if not treated promptly \cite{khan2025diagnostic}. In the initial stages of DR, retinal capillaries are damaged due to hyperglycemia, which weakens the capillary walls and leads to Microaneurysms (MAs). MAs are small outpouchings in the lumen of the retinal vessels, typically measuring 15-60 $\mu m$ in diameter. Identification of MAs allows for timely recognition of DR, thus providing an opportunity for early intervention in patients \cite{arrigo2024digital}. To analyze them, fundus images are widely used \cite{mayya2021automated} where small red dots are an indication of MAs \cite{raghu2019transfusion}. Nevertheless, as shown in Fig. \ref{data}, MAs are tiny and inconspicuous with variations in brightness, contrast, and shape, making it difficult for physicians to detect them \cite{wu2024microseg}. Therefore, automated MA detection methods with high accuracy in fundus images are of great significance.

To achieve this goal, different methods are proposed, among which most are discriminative models such as segmentation models \cite{xu2024hacdr, jiang2024glanceseg, foo2023multi, yap2023cut}. In contrast to classification models lacking specific localization capability, segmentation models can provide detailed information about the MA boundaries, thus helping to assess the severity of lesions and enhance the interpretability of image-level information \cite{jiang2023eye}. Nevertheless, the challenges of data annotation and segmentation accuracy restrict the development of these methods. First, the tiny size of MAs and their morphological similarity to normal vascular structures create significant inter-observer variability in manual annotations \cite{mayya2021automated}. This label ambiguity propagates through supervised segmentation frameworks, leading to suboptimal boundary delineation. Second, the class imbalance problem is exacerbated in MA diagnosis, where positive pixels constitute less than 1\% of total image area in early-stage DR cases \cite{porwal2020idrid}. Traditional segmentation methods tend to converge to trivial solutions that ignore subtle MA features.

% Unlike discriminative models, generative models have been gradually applied to reconstruction-based methods for medical anomaly detection, mainly based on Auto-Encoders (AEs) \cite{cai2024rethinking, kascenas2023role, zhao2023ae}, Generative Adversarial Networks (GANs) \cite{bougaham2025industrial, xiang2023squid, schlegl2019f} and diffusion models \cite{kumar2025self, bercea2024diffusion, behrendt2024patched, wyatt2022anoddpm, wolleb2022diffusion}. By identifying deviations from normal patterns as anomalies, reconstruction-based methods can effectively detect small, irregular lesions while reducing reliance on large numbers of pixel-level annotations. Recently, diffusion-based methods have made progress towards more accurate anomaly localization under weak or no supervision by iteratively improving the capture of fine-grained lesion details, holding significant potential for MA detection. Nonetheless, utilizing diffusion models for reconstruction-based MA detection faces the following challenges: 

% \begin{figure}[!t]
% \centering
% \includegraphics[width=0.9\columnwidth]{para_pdf.pdf}
% \caption{Comparison of paradigms for diffusion-based medical AD: (a) noise-addition-denoising, and (b) image-conditioned denoising (ours).}
% \label{para}
% \end{figure}

Unlike discriminative models, generative models have been gradually applied to reconstruction-based methods for medical Anomaly Detection (AD), mainly based on Auto-Encoders (AEs), Generative Adversarial Networks (GANs) \cite{goodfellow2014generative} and diffusion models \cite{ho2020denoising, ma2024followyouremoji, ma2025followfaster, ma2025followyourclick, sun2025gl, shao2025trace}. By identifying deviations from normal patterns as anomalies, these methods can effectively detect small, irregular lesions while reducing reliance on large numbers of accurate pixel-level annotations. Recently, diffusion-based methods \cite{kumar2025self, fontanella2024diffusion, wyatt2022anoddpm, wolleb2022diffusion} have made progress towards more accurate anomaly localization by iteratively improving the capture of fine-grained lesion details \cite{sun2025bs}, holding significant potential for MA detection. Nonetheless, the following challenges still remain:

\begin{itemize}

    \item The inherent risk of learning ``\textbf{identity mapping}" still persists in existing frameworks based on diffusion models. ``Identity mapping" refers to the behavior of directly copying the input as output, whether normal or abnormal \cite{guo2025dinomaly}. This contradicts the foundational assumption that anomalies induce significant reconstruction deviations, ultimately causing false negatives.

    \item The inability to distinguish MAs from other anomalies leads to \textbf{high false positives}. Existing methods lacking pixel-level supervision signals tend to treat all reconstruction errors as homogeneous indicators of abnormality, disregarding the unique morphological and contextual signatures of the target anomalies. Consequently, confounding factors such as imaging artifacts or coexisting lesions can be indiscriminately flagged as MA candidates, undermining clinical utility.

    \item The \textbf{suboptimal reconstruction quality} of normal features hampers the performance of AD. In retinal imaging, incomplete restoration of vascular patterns may introduce spurious reconstruction errors, masking true MA lesions or misclassifying normal variations as anomalies. 
    
\end{itemize}

% As illustrated in Fig. \ref{para}, e
Existing diffusion-based methods mitigate ``identity mapping" through noise-addition-denoising \cite{kumar2025self, fontanella2024diffusion, li2024vague, wyatt2022anoddpm} in the inference phase. This strategy faces a fundamental resolution conflict: MAs and fine vascular details occupy overlapping high-frequency bands, yet demand diametrically opposed noise treatments. Reliable MA suppression requires near-complete high-frequency erosion, while precise vasculature reconstruction necessitates preserving those exact frequency components. Insufficient noise preserves anomalies while excessive noise obliterates details. Consequently, single noise calibration during inference becomes intrinsically paradoxical for these methods. 

Furthermore, the absence of pixel-level supervision elevates false positive rates. These models erroneously classify imaging artifacts and non-MA pathologies as MAs, which is clinically unacceptable. To introduce pixel-level supervision signals, self-supervised image-conditioned approaches like Img-Cond \cite{baugh2024image} have been proposed. However, unprocessed input conditioning propagates anomalies via ``identity mapping", while the spatial distributional bias introduced by synthetic anomalies could further diminish the performance. Notably, although pixel-level supervision has recently been demonstrated to boost AD performance in complex industrial scenarios \cite{baitieva2024supervised}, analogous exploration and validation remain scarce in the medical field. The intrinsic spatial linkage between lesions and their anatomical context \cite{shao2025rethinking} poses unique challenges for effective supervision, underscoring the need for tailored strategies.
% \begin{figure}[t]
% \centering
% \includegraphics[width=0.9\columnwidth]{method_comp_pdf.pdf}
% \caption{Overview of four paradigms for diffusion-based AD: (a) unsupervised/weakly supervised unconditional AD, (b) self-supervised conditional AD, (c) vanilla supervised conditional AD, and (d) our supervised conditional AD. Red dots in fundus images represents virtual MA anomalies.}
% \label{method_comp}
% \end{figure}

\begin{figure*}[!th]
\centering
\includegraphics[width=0.95\textwidth]{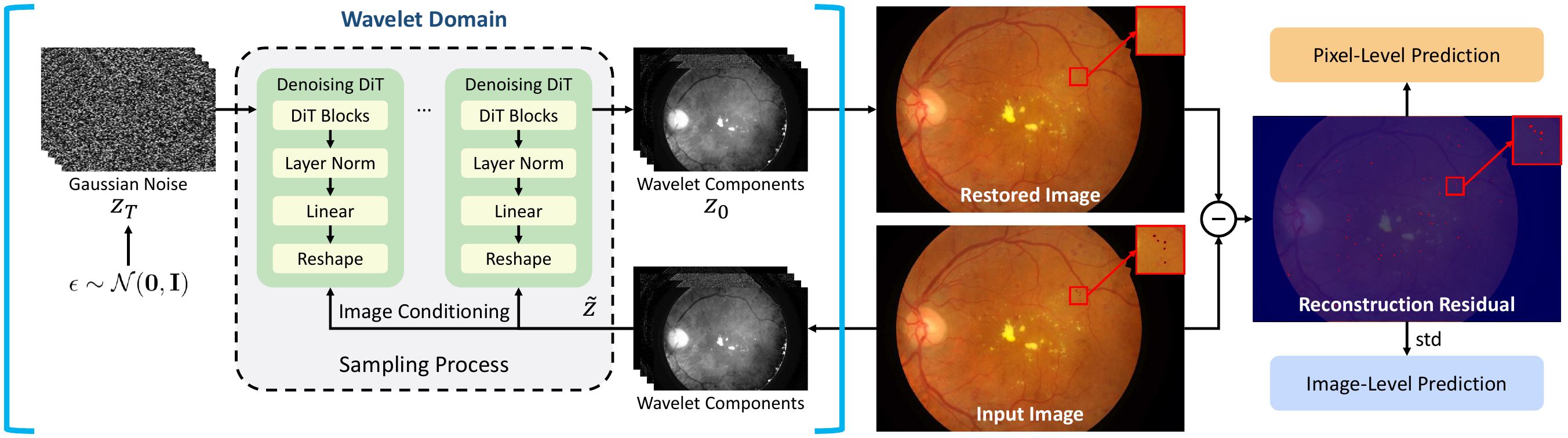}
\caption{Overview of our proposed WDT-MD method. It is a supervised DiT-based AD framework operating in the wavelet domain, which focuses on MA detection in fundus images. By synthesizing the normal pattern and subtracting the input from it, the model obtains an anomaly map, which further outputs both pixel-level and image-level predictions. (std: standard deviation)}
\label{frame_inference}
\end{figure*}

To address these challenges, we propose a Wavelet Diffusion Transformer framework for MA Detection (WDT-MD). This is a supervised image-conditioned wavelet-domain model based on  Diffusion Transformers (DiTs) \cite{peebles2023scalable, feng2025dit4edit}. Our contributions can be summarized as follows:

\begin{itemize}

    \item In order to mitigate ``identity mapping", we propose a \textbf{noise-encoded image conditioning} mechanism for diffusion-based MA detection. By perturbing the image condition with random intensities during training, the model is driven to capture the normal pattern. 

    \item To alleviate the issue of high false positives, we introduce pixel-level supervision signals in the training process through \textbf{pseudo-normal pattern synthesis}. Specifically, we obtain the pseudo-normal labels align with the spatial distribution of real fundus images using inpainting techniques. This enables the model to distinguish MAs from other anomalies, thereby improving the detection performance.

    \item To improve the reconstruction quality of normal features, we propose a \textbf{wavelet diffusion Transformer} architecture, which combines the global modelling capability of DiTs with the multi-scale analysis advantage of wavelet decomposition to better understand the overall structure and detailed information of fundus images.

    \item Comprehensive experiments on the IDRiD and e-ophtha MA datasets demonstrate exceptional performance of our WDT-MD, holding significant promise for improving early DR screening.
    
\end{itemize}

\section{Method}

We propose WDT-MD, a novel Wavelet Diffusion Transformer framework for MA Detection in fundus images, as illustrated in Fig. \ref{frame_inference}. This framework addresses key limitations of existing diffusion-based AD approaches for identifying MAs, which are critical early indicators of DR. The WDT-MD method initiates sampling from Gaussian noise in the wavelet domain, conditioned on the input fundus data. Through $T$-step iterative sampling, it reconstructs pseudo-normal fundus data. Subsequently, in the pixel domain, the reconstruction residual is computed by subtracting the input image from the restored pseudo-normal image. This residual map is then processed to yield both pixel-level segmentation and image-level classification predictions.

\subsection{Wavelet Diffusion Transformer}

Accurate reconstruction of normal retinal features is critical for reliable MA detection in DR screening. Existing Transformer-based backbones like U-ViT \cite{bao2023all} offer powerful global modeling capabilities for diffusion models, enabling them to better capture contextual information such as the spatial distribution of MAs. Nevertheless, their operation directly in the pixel space presents limitations. These approaches struggle to capture and preserve the intricate multi-scale structural and textural details inherent in retinal vasculature, and incur significant computational costs. To address these issues, subsequent works \cite{peebles2023scalable, ma2024sit, esser2024scaling} migrate the diffusion process into a learnable latent space using AE-based tokenizers. However, this two-stage strategy introduces its own challenges: the computational overhead of the tokenizer itself and the potential risk of losing inconspicuous features during the encoding or decoding process. This information loss is particularly problematic for detecting MAs. MAs, often subtle in size, demand a representation that inherently separates low-frequency contextual information such as vessel structures and backgrounds from high-frequency details including tiny lesions and textures.

\begin{figure}[!t]
\centering
\includegraphics[width=0.95\columnwidth]{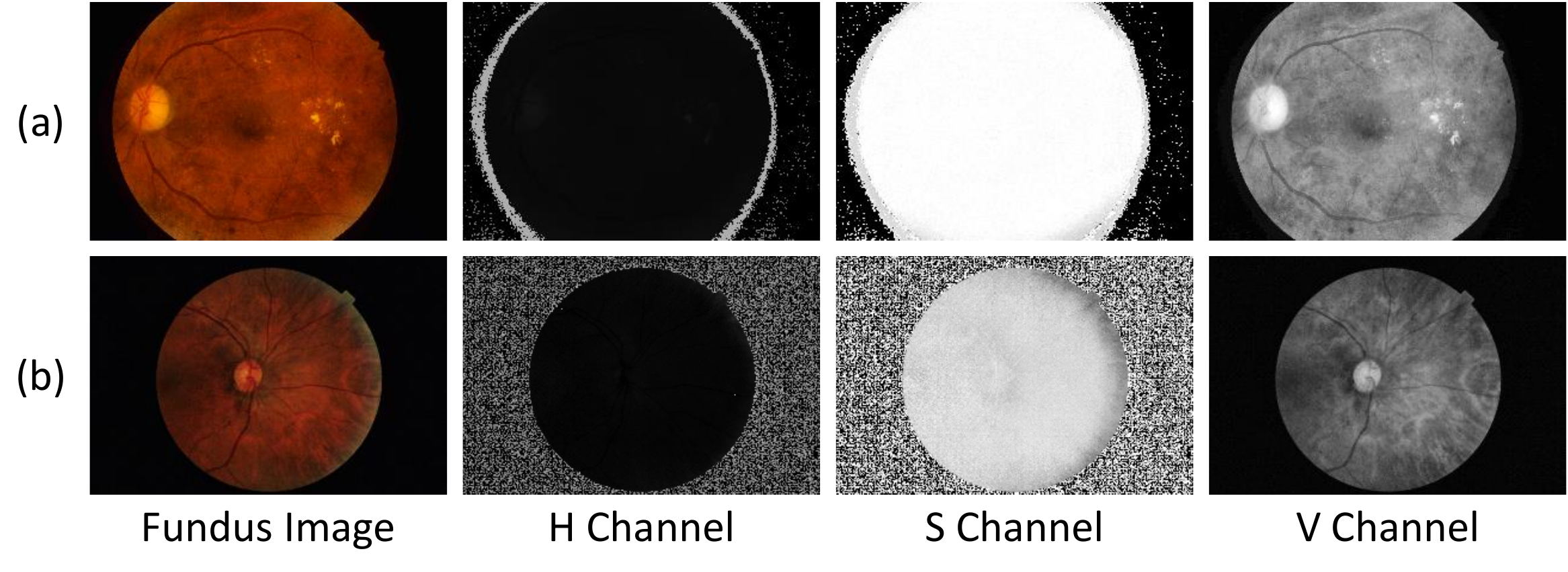}
\caption{Visualization of HSV channel decomposition on (a) IDRiD and (b) e-ophtha MA. The H and S channels carry little effective information but notable noise, while V contains almost all crucial structural and textural features.}
\label{hsv}
\end{figure}

To overcome these limitations as well as inspired by the success of wavelet analysis in low-level vision tasks \cite{zhao2024wavelet, huang2024wavedm}, we integrate Discrete Wavelet Transformation (DWT) with DiTs, proposing a wavelet diffusion Transformer architecture. Compared to AE-based tokenizers, DWT exhibits near-lossless reconstruction capabilities \cite{wang2024learning} and incurs lower computational overhead. Specifically, for an image \(I \in \mathbb{R}^{C \times H \times W}\), DWT transforms its Value channel \(V = max(R,G,B)\in \mathbb{R}^{1 \times H \times W}\) into four sub-bands:
\begin{equation}
V_{LL},\{V_{LH},V_{HL},V_{HH}\}=DWT(V),
\label{sub-bands}
\end{equation}
where $V_{L L},\left\{V_{L H}, V_{H L}, V_{H H}\right\}$ denote the low-frequency component of the image and high-frequency components in the vertical, horizontal, and diagonal directions, respectively. The selection of V channel helps alleviate the interference of imaging noise while effectively reducing the computational load, as illustrated in Fig. \ref{hsv}. Subsequently, the sub-bands are concatenated together, denoted as $z$:
\begin{equation}
z=Concat(V_{LL},V_{LH},V_{HL},V_{HH}).
\label{concat}
\end{equation}

Our wavelet diffusion Transformer incorporates both a forward process and a reverse process in the wavelet domain. The forward process can be defined as:
\begin{equation}
z_t=\sqrt{\bar{\alpha}_t} z_0+\sqrt{1-\bar{\alpha}_t} \epsilon, \quad \epsilon \sim \mathcal{N}(\mathbf{0}, \mathbf{I}),
\label{eq_noise}
\end{equation}
where $z_t$ is the wavelet components at timestep $t$, $\epsilon$ is a Gaussian noise map, and $\bar{\alpha}_t:=\prod_{s=0}^t \alpha_s$. Here, $\alpha_t=1-\beta_t$ is a differentiable function of timestep $t$. The diffusion loss is expressed as:
\begin{equation}
\mathcal{L}\left(\epsilon_\theta\right)=\sum_{t=1}^T \mathbb{E}_{z_0, \epsilon}\left[\left\|\epsilon_\theta\left(z_t, t, \tilde{z}\right)-\epsilon\right\|_2^2\right],
\label{eq_loss}
\end{equation}
where $\epsilon_\theta$ represents the predicted noise at timestep $t$ by the denoising DiT with parameters $\theta$, $T$ is the total diffusion timesteps, and $\tilde{z}$ is a given image condition.

During the reverse process, starting from Gaussian noise $z_T \sim \mathcal{N}(\mathbf{0}, \mathbf{I})$, the original sample $z_0$ is predicted through a multi-step denoising process:
\begin{equation}
z_{t-1}= 
\begin{cases}
\sqrt{\alpha_{t-1}}\left(c_{\text {out}}(t) \hat{z_0}+c_{\text {skip}}(t) z_t\right)+ \gamma \epsilon, & \!\!1<t \leqslant T_s \\
c_{\text {out}}(t) \hat{z_0}+c_{\text {skip}}(t) z_t, & \!\!t = 1,
\end{cases}
\label{eq_inf}
\end{equation}
where $\hat{z_0}=\frac{z_t-\frac{1-\alpha_t}{\sqrt{1-\bar{\alpha}_t}} \epsilon_\theta\left(z_t, t, \tilde{z}\right)}{\sqrt{\alpha_t}}$ is the predicted original sample, $\gamma=\frac{1-\alpha_{t-1}}{\sqrt{1-\bar{\alpha}_{t-1}}}$ is the scaling factor for Gaussian noise $\epsilon$, $T_s$ is the total sampling timesteps, and both $c_{\text {out }}(t)$ and $c_{\text {skip }}(t)$ are differentiable with $c_{\text {out }}(0)=0$ and $c_{\text {skip }}(0)=1$.

Subsequently, $z_0$ is split by channel and transformed into restored Value channel $V_0$ via Inverse DWT (IDWT):
\begin{equation}
V_{0,LL},V_{0,LH},V_{0,HL},V_{0,HH}=Split(z_0),
\label{split}
\end{equation}
\begin{equation}
V_0=IDWT(V_{0,LL},\{V_{0,LH},V_{0,HL},V_{0,HH}\}).
\label{idwt}
\end{equation}
Finally, $V_0$ is merged with the Hue channel $H$ and Saturation channel $S$ of the input image to obtain the restored image $I_0$.

\begin{figure}[!t]
\centering
\includegraphics[width=\columnwidth]{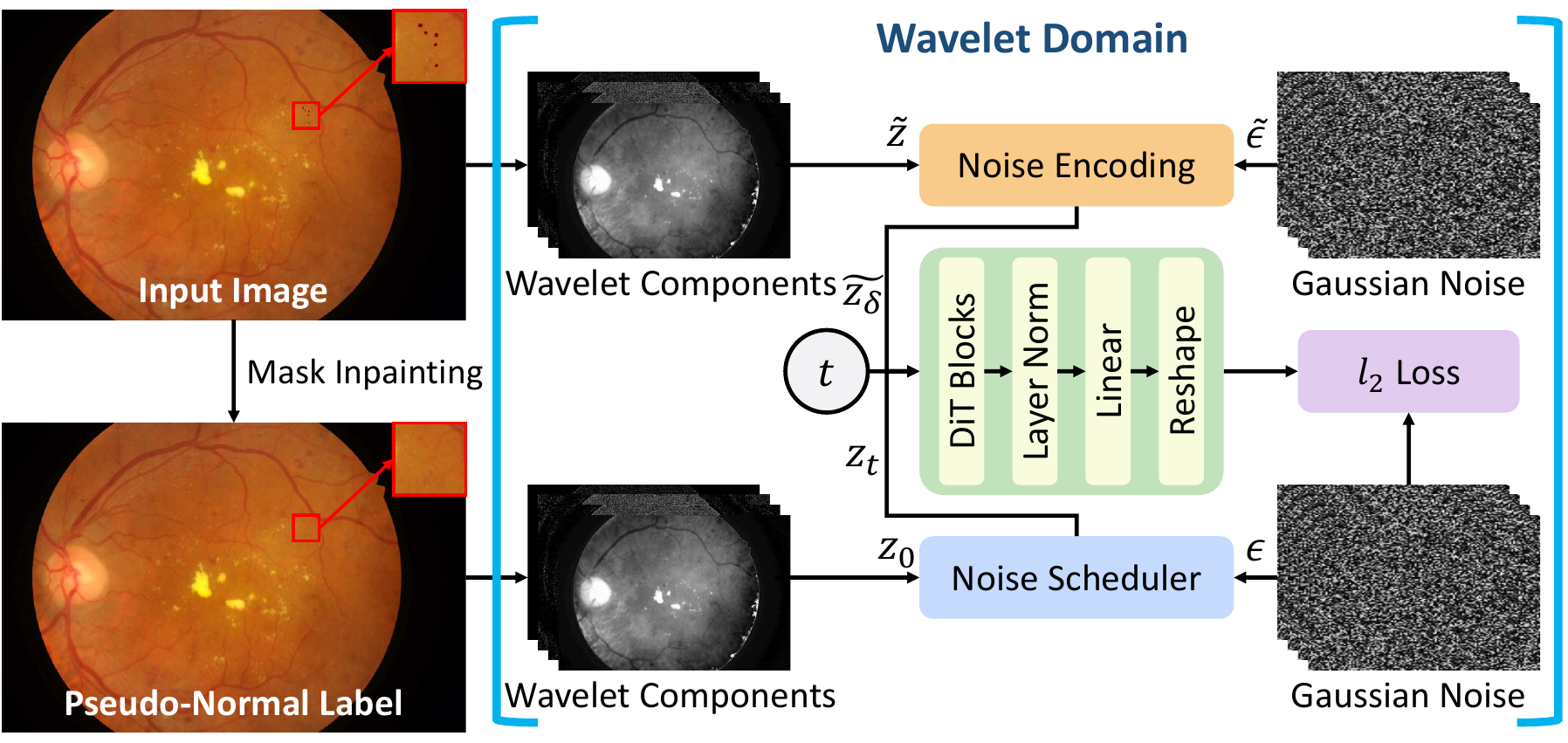}
\caption{The training process of our WDT-MD. }
\label{frame_training}
\end{figure}

\subsection{Noise-Encoded Image Conditioning}

To address the ``identity mapping" dilemma in diffusion models, we propose a noise-encoded image conditioning mechanism. In contrast to existing noise-addition-denoising methods that adopt single noise calibration during inference, we avoid ``identity mapping" by perturbing the image condition $\tilde{z}$ into noise-encoded $\widetilde{z_\delta}$ during training, denoted as:
\begin{equation}
\widetilde{z_\delta}=\sqrt{\bar{\alpha}_\delta} \tilde{z}+\sqrt{1-\bar{\alpha}_\delta} \epsilon, \quad \epsilon \sim \mathcal{N}(\mathbf{0}, \mathbf{I}),
\label{noi}
\end{equation}
where the added noise is determined by the timestep $\delta \in\{1,2, \ldots, \delta_{max}\}$ utilizing the noise scheduler. Correspondingly, the diffusion loss in Eq. (\ref{eq_loss}) becomes as follows:
\begin{equation}
\mathcal{L}\left(\epsilon_\theta\right)=\sum_{t=1}^T \mathbb{E}_{z_0, \delta, \epsilon}\left[\left\|\epsilon_\theta\left(z_t, t, \widetilde{z_\delta}\right)-\epsilon\right\|_2^2\right].
\label{eq_loss2}
\end{equation}
By introducing dynamic noise perturbations during training, this mechanism enforces the model to learn the underlying normal patterns of retinal structures rather than directly copying input pixels, as illustrated in Fig. \ref{frame_training}.

\begin{table*}[!p]

  \setlength{\tabcolsep}{2.6mm}
  \centering
  \begin{tabular}{cccccccccccc}
    \toprule

    \multirow{2.4}{*}{Method}     & \multirow{2.4}{*}{Source} & \multicolumn{5}{c}{Pixel-level} & \multicolumn{5}{c}{Image-level}\\
    \cmidrule(r){3-7}
    \cmidrule(r){8-12}
    & & AUC & ACC & F1 & SEN & SPE & AUC & ACC & F1 & SEN & SPE\\
    \midrule
    AnoDDPM & CVPR$_{22}$   & \cellcolor{third}{81.76}  & 99.91 & 53.34 & \cellcolor{third}{63.62} & 99.93 & \cellcolor{third}{71.90} & \cellcolor{third}{76.92} & 62.50 & \cellcolor{third}{55.56}  & \cellcolor{second}{88.24}  \\
    CPC & WACV$_{23}$   & 76.77  & \cellcolor{third}{99.93} & 49.75 & 53.63 & \cellcolor{third}{99.96} & \cellcolor{second}{77.45} & \cellcolor{second}{80.77} & \cellcolor{second}{70.59} & \cellcolor{second}{66.67}  & \cellcolor{second}{88.24}  \\
    DAE & MedIA$_{23}$   & 71.52  & 99.69 & 35.64 & 43.23 & 99.72 & 56.86 & 53.85 & 50.00 & \cellcolor{second}{66.67}  & 47.06  \\
    HACDR-Net & AAAI$_{24}$   & 56.38  & 95.07 & 4.03 & 18.82 & 95.12 & 63.07 & 65.38 & 52.63 & \cellcolor{third}{55.56}  & 70.59  \\
    AE[$d_{optimal}$] & MICCAI$_{24}$   & 75.06  & 99.24 & 19.52 & 50.88 & 99.27 & 62.75 & 61.54 & 54.55 & \cellcolor{second}{66.67}  & 58.82  \\
    Dif-fuse & TMI$_{24}$   & \cellcolor{second}{81.82}  & \cellcolor{second}{99.95} & \cellcolor{second}{69.55} & \cellcolor{second}{63.65} & \cellcolor{second}{99.97} & 71.57 & 73.08 & \cellcolor{third}{63.16} & \cellcolor{second}{66.67}  & \cellcolor{third}{76.47}  \\
    GatingAno & PR$_{24}$   & 78.73  & 92.07 & 11.49 & 63.04 & 92.09 & 54.25 & 53.85 & 45.45 & \cellcolor{third}{55.56}  & 52.94  \\
    DTU-Net & WACV$_{25}$   & 75.70  & \cellcolor{second}{99.95} & \cellcolor{third}{58.68} & 51.44 & \cellcolor{second}{99.97} & 68.63 & 69.23 & 60.00 & \cellcolor{second}{66.67}  & 70.59  \\
    \midrule
    WDT-MD & Ours   & \cellcolor{first}{\textbf{82.80}}  & \cellcolor{first}{\textbf{99.96}} & \cellcolor{first}{\textbf{74.43}} & \cellcolor{first}{\textbf{65.61}} & \cellcolor{first}{\textbf{99.98}} & \cellcolor{first}{\textbf{85.95}} & \cellcolor{first}{\textbf{88.46}} & \cellcolor{first}{\textbf{82.35}} & \cellcolor{first}{\textbf{77.78}}  & \cellcolor{first}{\textbf{94.12}}  \\
    
    \bottomrule
    % \multicolumn{12}{l}{\small $\star$ Segmentation methods.}\\
  \end{tabular}
    \caption{Quantitative comparison of the proposed WDT-MD method with the state-of-the-art methods on the IDRiD dataset. Best results are highlighted as \colorbox{first}{\textbf{first}}, \colorbox{second}{second} and \colorbox{third}{third}. (Unit: \%)}
      \label{comp_tab1}

\end{table*}

\begin{table*}[!p]

  \setlength{\tabcolsep}{2.6mm}
  \centering
  \begin{tabular}{cccccccccccc}
    \toprule

    \multirow{2.4}{*}{Method}     & \multirow{2.4}{*}{Source} & \multicolumn{5}{c}{Pixel-level} & \multicolumn{5}{c}{Image-level}\\
    \cmidrule(r){3-7}
    \cmidrule(r){8-12}
    & & AUC & ACC & F1 & SEN & SPE & AUC & ACC & F1 & SEN & SPE\\
    \midrule
    AnoDDPM & CVPR$_{22}$   & 79.26  & \cellcolor{third}{99.96} & 32.09 & 58.56 & \cellcolor{third}{99.97} & 60.00 & 61.54 & 51.61 & \cellcolor{third}{53.33}  & \cellcolor{third}{66.67}  \\
    CPC & WACV$_{23}$   & 76.28  & \cellcolor{second}{99.98} & \cellcolor{third}{35.34} & 52.57 & \cellcolor{second}{99.98} & \cellcolor{second}{65.42} & \cellcolor{second}{66.67} & \cellcolor{second}{58.06} & \cellcolor{second}{60.00}  & \cellcolor{second}{70.83}  \\
    DAE & MedIA$_{23}$   & 72.64  & 99.95 & 21.96 & 45.31 & 99.96 & 57.08 & 56.41 & 51.43 & \cellcolor{second}{60.00}  & 54.17  \\
    HACDR-Net & AAAI$_{24}$   & 54.13  & 98.03 & 3.53 & 9.56 & 98.04 & 41.67 & 43.59 & 31.25 & 33.33  & 50.00  \\
    AE[$d_{optimal}$] & MICCAI$_{24}$   & 78.86  & \cellcolor{second}{99.98} & 32.21 & 57.75 & \cellcolor{second}{99.98} & 62.08 & \cellcolor{third}{64.10} & 53.33 & \cellcolor{third}{53.33}  & \cellcolor{second}{70.83}  \\
    Dif-fuse & TMI$_{24}$   & \cellcolor{second}{80.82}  & \cellcolor{third}{99.96} & 32.48 & \cellcolor{second}{61.67} & 99.96 & 61.25 & 61.54 & 54.55 & \cellcolor{second}{60.00}  & 62.50  \\
    GatingAno & PR$_{24}$   & 78.27  & 98.83 & 2.253 & 58.45 & 98.84 & \cellcolor{third}{63.33} & \cellcolor{third}{64.10} & \cellcolor{third}{56.25} & \cellcolor{second}{60.00}  & \cellcolor{third}{66.67}  \\
    DTU-Net & WACV$_{25}$   & \cellcolor{third}{80.72}  & \cellcolor{second}{99.98} & \cellcolor{second}{42.99} & \cellcolor{third}{61.46} & \cellcolor{second}{99.98} & 49.58 & 48.72 & 44.44 & \cellcolor{third}{53.33}  & 45.83  \\

    \midrule
    WDT-MD & Ours   & \cellcolor{first}{\textbf{81.08}}  & \cellcolor{first}{\textbf{99.99}} & \cellcolor{first}{\textbf{57.70}} & \cellcolor{first}{\textbf{62.16}} & \cellcolor{first}{\textbf{99.99}} & \cellcolor{first}{\textbf{70.83}} & \cellcolor{first}{\textbf{71.79}} & \cellcolor{first}{\textbf{64.52}} & \cellcolor{first}{\textbf{66.67}}  & \cellcolor{first}{\textbf{75.00}}  \\
    
    \bottomrule
    % \multicolumn{12}{l}{\small $\star$ Segmentation methods.}\\

  \end{tabular}
    \caption{Quantitative comparison of the proposed WDT-MD method with the state-of-the-art methods on the e-ophtha MA dataset. Best results are highlighted as \colorbox{first}{\textbf{first}}, \colorbox{second}{second} and \colorbox{third}{third}. (Unit: \%)}
      \label{comp_tab2}

\end{table*}

\subsection{Pseudo-Normal Pattern Synthesis}

To mitigate false positives, we synthesize pseudo-normal labels using inpainting techniques, thus introducing pixel-level supervision signals in the training process, which can be briefly expressed as follows:
\begin{equation}
V_{pn}=(1-M) \odot V+M \odot \mathcal{I}(V, M),
\label{eq_loss2}
\end{equation}
where $V_{pn}$ represents the pseudo-normal V channel, $M$ denotes the binary MA mask normalized to $[0,1]$, and $\mathcal{I}$ is the inpainting algorithm. Here, we employ Telea, a classical inpainting method \cite{telea2004image}. In contrast to synthesizing anomalies on normal fundus images \cite{sun2025unseen}, our core idea is to infer unknown pseudo-normal regions from known normal pixels guided by MA masks, ensuring the spatial distribution accuracy of pixel-level supervision.

\begin{table*}[!p]

  \setlength{\tabcolsep}{3.2mm}
  \centering
  \begin{tabular}{>{\centering}p{1cm}>{\centering}p{1cm}cccccccccc}
    \toprule

    \multirow{2.4}{*}{$\tau$} & \multirow{2.4}{*}{$\psi$} & \multicolumn{5}{c}{Pixel-level} & \multicolumn{5}{c}{Image-level}\\
    \cmidrule(r){3-7}
    \cmidrule(r){8-12}
    & & AUC & ACC & F1 & SEN & SPE & AUC & ACC & F1 & SEN & SPE\\
    \midrule
        
    \XSolidBrush & \XSolidBrush   & \cellcolor{second}{73.17}  & \cellcolor{third}{97.17} & \cellcolor{second}{6.39} & \cellcolor{third}{49.28} & \cellcolor{third}{97.20} & \cellcolor{second}{48.37} & \cellcolor{second}{46.15} & \cellcolor{second}{41.67} & \cellcolor{third}{55.56}  & \cellcolor{second}{41.18}  \\
    
    \XSolidBrush & \Checkmark   & 49.98  & \cellcolor{second}{98.78} & 0.53 & 0.30 & \cellcolor{second}{98.84} & \cellcolor{third}{42.81} & \cellcolor{third}{42.31} & 34.78 & 44.44  & \cellcolor{second}{41.18}  \\

    \Checkmark & \XSolidBrush   & \cellcolor{third}{67.20}  & 68.84 & \cellcolor{third}{0.87} & \cellcolor{second}{62.60} & 68.84 & 42.16 & 34.62 & \cellcolor{third}{41.38} & \cellcolor{second}{66.67}  & \cellcolor{third}{17.65}  \\

    \Checkmark & \Checkmark   & \cellcolor{first}{\textbf{82.80}}  & \cellcolor{first}{\textbf{99.96}} & \cellcolor{first}{\textbf{74.43}} & \cellcolor{first}{\textbf{65.61}} & \cellcolor{first}{\textbf{99.98}} & \cellcolor{first}{\textbf{85.95}} & \cellcolor{first}{\textbf{88.46}} & \cellcolor{first}{\textbf{82.35}} & \cellcolor{first}{\textbf{77.78}}  & \cellcolor{first}{\textbf{94.12}}   \\
    
    \bottomrule

  \end{tabular}
    \caption{Ablation study of core components of WDT-MD on IDRiD. $\tau$ denotes noise encoding in image conditioning, and $\psi$ denotes pixel-level supervision. Best results are highlighted as \colorbox{first}{\textbf{first}}, \colorbox{second}{second} and \colorbox{third}{third}. (Unit: \%)}
      \label{abl1}

\end{table*}

\begin{table*}[!p]

  \setlength{\tabcolsep}{3.2mm}
  \centering
  \begin{tabular}{>{\centering}p{1cm}>{\centering}p{1cm}cccccccccc}
    \toprule

    \multirow{2.4}{*}{$\tau$} & \multirow{2.4}{*}{$\psi$} & \multicolumn{5}{c}{Pixel-level} & \multicolumn{5}{c}{Image-level}\\
    \cmidrule(r){3-7}
    \cmidrule(r){8-12}
    & & AUC & ACC & F1 & SEN & SPE & AUC & ACC & F1 & SEN & SPE\\
    \midrule
        
    \XSolidBrush & \XSolidBrush   & \cellcolor{third}{58.78}  & \cellcolor{third}{97.03} & \cellcolor{third}{0.36} & \cellcolor{third}{20.47} & \cellcolor{third}{97.04} & \cellcolor{third}{47.08} & \cellcolor{third}{48.72} & \cellcolor{third}{37.50} & \cellcolor{third}{40.00}  & 54.17  \\
    
    \XSolidBrush & \Checkmark   & 51.69  & \cellcolor{second}{98.14} & \cellcolor{second}{5.29} & 4.05 & \cellcolor{second}{98.15} & 45.83 & \cellcolor{third}{48.72} & 33.33 & 33.33  & \cellcolor{third}{58.33}  \\

    \Checkmark & \XSolidBrush   & \cellcolor{second}{68.06}  & 79.84 & 0.17 & \cellcolor{second}{56.97} & 79.84 & \cellcolor{second}{61.25} & \cellcolor{second}{61.54} & \cellcolor{second}{54.55} & \cellcolor{second}{60.00}  & \cellcolor{second}{62.50}  \\

    \Checkmark & \Checkmark   & \cellcolor{first}{\textbf{81.08}}  & \cellcolor{first}{\textbf{99.99}} & \cellcolor{first}{\textbf{57.70}} & \cellcolor{first}{\textbf{62.16}} & \cellcolor{first}{\textbf{99.99}} & \cellcolor{first}{\textbf{70.83}} & \cellcolor{first}{\textbf{71.79}} & \cellcolor{first}{\textbf{64.52}} & \cellcolor{first}{\textbf{66.67}}  & \cellcolor{first}{\textbf{75.00}}  \\
    
    \bottomrule
    % \multicolumn{12}{l}{\small $\star$ Segmentation methods.}\\

  \end{tabular}
    \caption{Ablation study of core components of WDT-MD on e-ophtha MA. $\tau$ denotes noise encoding in image conditioning, and $\psi$ denotes pixel-level supervision. Best results are highlighted as \colorbox{first}{\textbf{first}}, \colorbox{second}{second} and \colorbox{third}{third}. (Unit: \%)}
      \label{abl2}

\end{table*}

\begin{table*}[!p]

  \setlength{\tabcolsep}{1.7mm}
  \centering
  \begin{tabular}{ccccccccccccc}
    \toprule

    \multirow{2.4}{*}{Tokenizer}     & \multicolumn{5}{c}{Pixel-level} & \multicolumn{5}{c}{Image-level} &\multirow{2.4}{*}{Params (M)} &\multirow{2.4}{*}{FLOPs (G)}\\
    \cmidrule(r){2-6}
    \cmidrule(r){7-11}
    & AUC & ACC & F1 & SEN & SPE & AUC & ACC & F1 & SEN & SPE\\
    \midrule

    \XSolidBrush   & 81.03  & 99.92 & 64.24 & 62.09 & 99.94  & \cellcolor{second}{77.45} & \cellcolor{second}{80.77} & \cellcolor{second}{70.59} & \cellcolor{second}{66.67} & \cellcolor{second}{88.24} & 41.49 & 478.30 \\
    
    AE-KL   & 80.81  & \cellcolor{third}{99.93} & 67.53 & 61.65 & \cellcolor{third}{99.95}  & 68.95 & \cellcolor{third}{73.08} & 58.82 & \cellcolor{third}{55.56} & \cellcolor{third}{82.35} & \cellcolor{second}{36.04} & 229.36 \\

    VQ-VAE   & \cellcolor{third}{81.31}  & 99.91 & \cellcolor{third}{68.19} & \cellcolor{third}{62.65} & 99.93  & \cellcolor{third}{71.57} & \cellcolor{third}{73.08} & \cellcolor{third}{63.16} & \cellcolor{second}{66.67} & 76.47 & \cellcolor{third}{36.24} & \cellcolor{second}{146.75} \\
    
    VQGAN   & \cellcolor{second}{81.37}  & \cellcolor{second}{99.95} & \cellcolor{second}{69.82} & \cellcolor{second}{62.76} & \cellcolor{second}{99.97}  & \cellcolor{second}{77.45} & \cellcolor{second}{80.77} & \cellcolor{second}{70.59} & \cellcolor{second}{66.67} & \cellcolor{second}{88.24} & 39.00 & \cellcolor{third}{150.96} \\
    
    \midrule

    DWT (Ours)   & \cellcolor{first}{\textbf{82.80}}  & \cellcolor{first}{\textbf{99.96}} & \cellcolor{first}{\textbf{74.43}} & \cellcolor{first}{\textbf{65.61}} & \cellcolor{first}{\textbf{99.98}} & \cellcolor{first}{\textbf{85.95}} & \cellcolor{first}{\textbf{88.46}} & \cellcolor{first}{\textbf{82.35}} & \cellcolor{first}{\textbf{77.78}}  & \cellcolor{first}{\textbf{94.12}}  & \cellcolor{first}{\textbf{35.04}}  & \cellcolor{first}{\textbf{119.76}} \\

    \bottomrule

  \end{tabular}
  \caption{Impact of various tokenizers for compression in WDT-MD on the IDRiD dataset. Best results are highlighted as \colorbox{first}{\textbf{first}}, \colorbox{second}{second} and \colorbox{third}{third}. (Unit: \%, Params: number of model parameters)}
    \label{abl3}

\end{table*}

\begin{table*}[!p]

  \setlength{\tabcolsep}{1.7mm}
  \centering
  \begin{tabular}{ccccccccccccc}
    \toprule

    \multirow{2.4}{*}{Tokenizer}     & \multicolumn{5}{c}{Pixel-level} & \multicolumn{5}{c}{Image-level} &\multirow{2.4}{*}{Params (M)} &\multirow{2.4}{*}{FLOPs (G)}\\
    \cmidrule(r){2-6}
    \cmidrule(r){7-11}
    & AUC & ACC & F1 & SEN & SPE & AUC & ACC & F1 & SEN & SPE\\
    \midrule

    \XSolidBrush   & 77.81  & 99.90 & 19.86 & 55.70 & 99.91  & 44.17 & 43.59 & 38.89 & \cellcolor{third}{46.67} & 41.67 & 41.49 & 478.30 \\
    
    AE-KL   & 78.22  & \cellcolor{third}{99.94} & 22.52 & 56.50 & 99.94  & \cellcolor{third}{53.75} & \cellcolor{third}{53.85} & \cellcolor{third}{47.06} & \cellcolor{second}{53.33} & \cellcolor{third}{54.17} & \cellcolor{second}{36.04} & 229.36 \\

    VQ-VAE   & \cellcolor{third}{79.89}  & \cellcolor{third}{99.94} & \cellcolor{third}{27.49} & \cellcolor{third}{59.84} & \cellcolor{third}{99.95}  & \cellcolor{second}{55.83} & \cellcolor{second}{56.41} & \cellcolor{second}{48.48} & \cellcolor{second}{53.33} & \cellcolor{second}{58.33} & \cellcolor{third}{36.24} & \cellcolor{second}{146.75} \\
    
    VQGAN   & \cellcolor{second}{79.98}  & \cellcolor{second}{99.97} & \cellcolor{second}{41.02} & \cellcolor{second}{59.99} & \cellcolor{second}{99.97}  & \cellcolor{second}{55.83} & \cellcolor{second}{56.41} & \cellcolor{second}{48.48} & \cellcolor{second}{53.33} & \cellcolor{second}{58.33} & 39.00 & \cellcolor{third}{150.96} \\
    
    \midrule

    DWT (Ours)   & \cellcolor{first}{\textbf{81.08}}  & \cellcolor{first}{\textbf{99.99}} & \cellcolor{first}{\textbf{57.70}} & \cellcolor{first}{\textbf{62.16}} & \cellcolor{first}{\textbf{99.99}} & \cellcolor{first}{\textbf{70.83}} & \cellcolor{first}{\textbf{71.79}} & \cellcolor{first}{\textbf{64.52}} & \cellcolor{first}{\textbf{66.67}}  & \cellcolor{first}{\textbf{75.00}} & \cellcolor{first}{\textbf{35.04}}  & \cellcolor{first}{\textbf{119.76}} \\

    \bottomrule

  \end{tabular}
    \caption{Impact of various tokenizers for compression in WDT-MD on the e-ophtha MA dataset. Best results are highlighted as \colorbox{first}{\textbf{first}}, \colorbox{second}{second} and \colorbox{third}{third}. (Unit: \%, Params: number of model parameters)}
      \label{abl4}

\end{table*}

\begin{table*}[!p]

  \setlength{\tabcolsep}{1.5mm}
  \centering
  \begin{tabular}{ccccccccccccc}
    \toprule

    \multirow{2.4}{*}{Backbone}     & \multicolumn{5}{c}{Pixel-level} & \multicolumn{5}{c}{Image-level} &\multirow{2.4}{*}{Params (M)} &\multirow{2.4}{*}{FLOPs (G)}\\
    \cmidrule(r){2-6}
    \cmidrule(r){7-11}
    & AUC & ACC & F1 & SEN & SPE & AUC & ACC & F1 & SEN & SPE\\
    \midrule

    Attention U-Net   & 79.88  & 99.86 & 56.32 & 59.85 & 99.88  & 71.57 & 73.08 & 63.16 & \cellcolor{second}{66.67} & 76.47 & 382.98 & 193.48 \\
    U-ViT   & 80.89  & \cellcolor{second}{99.94} & \cellcolor{second}{69.79} & 61.81 & \cellcolor{third}{99.96}  & 74.51 & 76.92 & 66.67 & \cellcolor{second}{66.67} & \cellcolor{third}{82.35} & 59.67 & 313.89 \\
    DiT (N = 1)   & 79.18  & \cellcolor{third}{99.93} & 61.31 & 58.40 & 99.95  & 63.07 & 65.38 & 52.63 & \cellcolor{third}{55.56} & 70.59 & \cellcolor{first}{\textbf{5.78}} & \cellcolor{first}{\textbf{10.17}} \\
    DiT (N = 2)   & 80.62  & 99.92 & 61.95 & 61.29 & 99.95  & 71.57 & 73.08 & 63.16 & \cellcolor{second}{66.67} & 76.47 & \cellcolor{second}{8.44} & \cellcolor{second}{20.14} \\
    DiT (N = 4)   & \cellcolor{third}{81.22}  & \cellcolor{second}{99.94} & 67.76 & \cellcolor{third}{62.46} & \cellcolor{second}{99.97}  & \cellcolor{third}{80.07} & \cellcolor{third}{80.77} & \cellcolor{third}{73.68} & \cellcolor{first}{\textbf{77.78}} & \cellcolor{third}{82.35} & \cellcolor{third}{13.76} & \cellcolor{third}{40.06} \\
    DiT (N = 8)   & \cellcolor{second}{81.93}  & \cellcolor{second}{99.94} & \cellcolor{third}{69.43} & \cellcolor{second}{63.90} & \cellcolor{second}{99.97}  & \cellcolor{second}{83.01} & \cellcolor{second}{84.62} & \cellcolor{second}{77.78} & \cellcolor{first}{\textbf{77.78}} & \cellcolor{second}{88.24} & 24.40 & 79.91 \\
    \midrule
    DiT (N = 12, Ours)   & \cellcolor{first}{\textbf{82.80}}  & \cellcolor{first}{\textbf{99.96}} & \cellcolor{first}{\textbf{74.43}} & \cellcolor{first}{\textbf{65.61}} & \cellcolor{first}{\textbf{99.98}} & \cellcolor{first}{\textbf{85.95}} & \cellcolor{first}{\textbf{88.46}} & \cellcolor{first}{\textbf{82.35}} & \cellcolor{first}{\textbf{77.78}}  & \cellcolor{first}{\textbf{94.12}}  & 35.04  & 119.76 \\
    % DiT (N = 16)   & 0  & 0 & 0 & 0 & 0  & 0 & 0 & 0 & 0 & 0 & 0 & 0 \\

    \bottomrule

  \end{tabular}
    \caption{Impact of various backbones of the noise estimator network in WDT-MD on the IDRiD dataset. Best results are highlighted as \colorbox{first}{\textbf{first}}, \colorbox{second}{second} and \colorbox{third}{third}. (N: number of DiT blocks, Unit: \%, Params: number of model parameters)}
      \label{abl5}

\end{table*}

\begin{table*}[!p]

  \setlength{\tabcolsep}{1.5mm}
  \centering
  \begin{tabular}{ccccccccccccc}
    \toprule

    \multirow{2.4}{*}{Backbone}     & \multicolumn{5}{c}{Pixel-level} & \multicolumn{5}{c}{Image-level} &\multirow{2.4}{*}{Params (M)} &\multirow{2.4}{*}{FLOPs (G)}\\
    \cmidrule(r){2-6}
    \cmidrule(r){7-11}
    & AUC & ACC & F1 & SEN & SPE & AUC & ACC & F1 & SEN & SPE\\
    \midrule

    Attention U-Net   & 79.70  & \cellcolor{third}{99.96} & 37.85 & 59.43 & 99.96  & 52.50 & 53.85 & 43.75 & 46.67 & \cellcolor{third}{58.33} & 382.98 & 193.48 \\
    U-ViT   & \cellcolor{second}{80.39}  & \cellcolor{third}{99.96} & 31.57 & \cellcolor{second}{60.83} & 99.96  & \cellcolor{second}{61.25} & \cellcolor{second}{61.54} & \cellcolor{second}{54.55} & \cellcolor{second}{60.00} & \cellcolor{second}{62.50} & 59.67 & 313.89 \\
    DiT (N = 1)   & 76.99  & \cellcolor{second}{99.97} & 33.90 & 54.02 & \cellcolor{third}{99.97}  & \cellcolor{third}{57.08} & \cellcolor{third}{56.41} & \cellcolor{third}{51.43} & \cellcolor{second}{60.00} & 54.17 & \cellcolor{first}{\textbf{5.78}} & \cellcolor{first}{\textbf{10.17}} \\
    DiT (N = 2)   & 77.41  & \cellcolor{second}{99.97} & 36.68 & 54.85 & \cellcolor{second}{99.98}  & 55.83 & \cellcolor{third}{56.41} & 48.48 & \cellcolor{third}{53.33} & \cellcolor{third}{58.33} & \cellcolor{second}{8.44} & \cellcolor{second}{20.14} \\
    DiT (N = 4)   & 79.97  & \cellcolor{second}{99.97} & \cellcolor{third}{44.37} & 59.96 & \cellcolor{second}{99.98}  & 51.67 & 51.28 & 45.71 & \cellcolor{third}{53.33} & 50.00 & \cellcolor{third}{13.76} & \cellcolor{third}{40.06} \\
    DiT (N = 8)   & \cellcolor{third}{80.10}  & \cellcolor{second}{99.97} & \cellcolor{second}{44.48} & \cellcolor{third}{60.21} & \cellcolor{second}{99.98}  & \cellcolor{third}{57.08} & \cellcolor{third}{56.41} & \cellcolor{third}{51.43} & \cellcolor{second}{60.00} & 54.17 & 24.40 & 79.91 \\
    \midrule
    DiT (N = 12, Ours)   & \cellcolor{first}{\textbf{81.08}}  & \cellcolor{first}{\textbf{99.99}} & \cellcolor{first}{\textbf{57.70}} & \cellcolor{first}{\textbf{62.16}} & \cellcolor{first}{\textbf{99.99}} & \cellcolor{first}{\textbf{70.83}} & \cellcolor{first}{\textbf{71.79}} & \cellcolor{first}{\textbf{64.52}} & \cellcolor{first}{\textbf{66.67}}  & \cellcolor{first}{\textbf{75.00}} & 35.04  & 119.76 \\
    % DiT (N = 16)   & 0  & 0 & 0 & 0 & 0  & 0 & 0 & 0 & 0 & 0 & 45.68 & 159.61 \\

    \bottomrule

  \end{tabular}
    \caption{Impact of various backbones of the noise estimator network in WDT-MD on the e-ophtha MA dataset. Best results are highlighted as \colorbox{first}{\textbf{first}}, \colorbox{second}{second} and \colorbox{third}{third}. (N: number of DiT blocks, Unit: \%, Params: number of model parameters)}
      \label{abl6}

\end{table*}

\section{Experiments}

\subsection{Datasets and Evaluation Metrics}
\subsubsection{Data Preparation.}

Two publicly available datasets, namely IDRiD \cite{porwal2018indian} and e-ophtha MA \cite{decenciere2013teleophta}, are adopted for extensive evaluation.

\textbf{The IDRiD dataset}, a benchmark resource for diabetic retinopathy analysis, was adapted for our study. For MA detection, we curated a subset of 249 samples, including 199 training cases, 24 validation cases, and 26 test cases. Specifically, the training set contains 134 normal images and 65 abnormal images. Contrast Limited Adaptive Histogram Equalization (CLAHE) was applied with 8 $\times$ 8 tile grids and a 2.0 clip limit to enhance contrast. Considering  the computational overhead, we implemented dimension standardization through bilinear downsampling to 300 $\times$ 200 pixels. 

\textbf{The e-ophtha MA dataset} consists of 381 cases divided into 304 training, 38 validation, and 39 test samples. Specifically, the training set contains 188 normal images and 116 abnormal images. The preprocessing pipeline maintained strict consistency with IDRiD: (1) CLAHE (8 $\times$ 8 tile grids, 2.0 clip limit); (2) downsampling to 300 $\times$ 200 pixels.

\subsubsection{Evaluation Metrics.}

To evaluate the MA detection performance, we calculate the Area Under Curve (AUC), Accuracy (ACC), F1 score, Sensitivity (SEN) and Specificity (SPE) for both pixel-level and image-level detection.

\begin{figure*}[!th]
\centering
\includegraphics[width=0.95\textwidth]{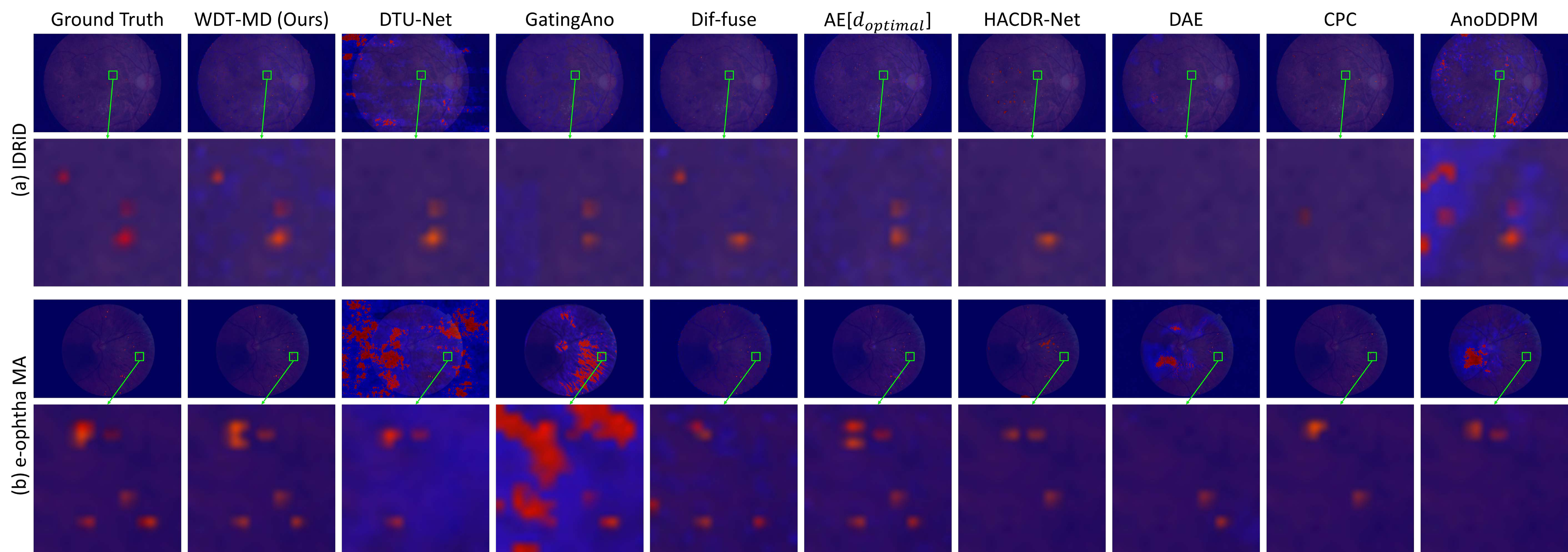}
\caption{The qualitative results for WDT-MD compared with other state-of-the-art methods on IDRiD and e-ophtha MA.}
\label{comp}
\end{figure*}

\subsection{Implementation Details}

All experiments were performed using PyTorch 2.5.1 on a single NVIDIA V100 32 GB GPU within Ubuntu 22.04. WDT-MD was trained from scratch over 600 epochs with a batch size of 4 utilizing the AdamW optimizer, complemented by a dynamic learning rate schedule initialized at $10^{-4}$. The noise scheduling parameter $\beta_t$ followed a scaled linear trajectory ranging from 0.00085 to 0.012 across $T=1000$ diffusion timesteps. The sampling steps $T_s$ was set to 50 using the LCM sampler \cite{luo2023latent}. In pseudo-normal pattern synthesis, the inpainting radius $r$ is set to 3 pixels. For wavelet decomposition, the Daubechies 6 basis was selected to balance computational efficiency and time-frequency localization \cite{wang2024wavelet}.

\subsection{Main Results}

On both the IDRiD and e-ophtha MA datasets, we benchmark our WDT-MD against other state-of-the-art methods: (1) diffusion-based AD methods DTU-Net \cite{kumar2025self}, Dif-fuse \cite{fontanella2024diffusion} and AnoDDPM \cite{wyatt2022anoddpm}; (2) a GAN-based AD method GatingAno \cite{zhang2024anomaly}; (3) AE-based AD methods AE[$d_{optimal}$] \cite{cai2024rethinking} and DAE \cite{kascenas2023role}; (4) U-Net-based segmentation methods HACDR-Net \cite{xu2024hacdr} and CPC \cite{yap2023cut}. 

Quantitative results are presented in Table \ref{comp_tab1} and Table \ref{comp_tab2}. WDT-MD demonstrates top performance, outperforming current state-of-the-art methods, spanning both AD and segmentation frameworks. At the pixel level, it achieves an AUC of 82.80\% on IDRiD and 81.08\% on e-ophtha MA. In terms of the image level, it reaches 85.95\% on IDRiD and 70.83\% on e-ophtha MA. Furthermore, Fig. \ref{comp} illustrates the qualitative results on both datasets, highlighting WDT-MD's remarkable ability to precisely detect subtle MAs.

\subsection{Ablation Study}

\subsubsection{Ablation Study of Core Components.}
Our ablation study highlights the notable improvements brought by our method, as shown in Table \ref{abl1} and Table \ref{abl2}. Specifically, the pixel-level SEN is improved to 65.61\% and 62.16\% on IDRiD and e-ophtha MA, respectively. This underscores that noise-encoded image conditioning effectively mitigates ``identity mapping" by preventing mere replication from the image condition. Furthermore, the integration of pixel-level supervision markedly reduces false positives, yielding improvements of (31.14\%/76.47\%) in (pixel-level/image-level) SPE on IDRiD and (20.15\%/12.50\%) on e-ophtha MA. 

\subsubsection{Impact of Tokenizers.}

To investigate the impact of different tokenization strategies, we conducted comparative experiments. As presented in Table \ref{abl3} and Table \ref{abl4}, DWT achieves the best performance, with improvements of (4.61\%/11.76\%) in (pixel-level/image-level) F1 score on IDRiD and (16.68\%/16.04\%) on e-ophtha MA. This underscores the advantage of DWT in detail preservation and multi-scale feature modeling. Notably, compared with training-based tokenizers such as VQGAN, the use of DWT reduces the Params by 1.00M and the FLOPs by 18.39\%, demonstrating its superiority in computational efficiency.

\subsubsection{Impact of Backbones.}

The impact of different backbones on WDT-MD performance is evident in Tables \ref{abl5} and \ref{abl6}. On both IDRiD and e-ophtha MA, our DiT backbone (N=12) delivers the best performance. Notably, compared with Attention U-Net \cite{dhariwal2021diffusion}, the most commonly used backbone in diffusion models, our backbone reduces the Params by 90.85\% and FLOPs by 38.10\%, reinforcing its suitability for clinical deployment.

\subsubsection{Impact of Noise Encoding Timesteps.}
\begin{figure}[!t]
\centering
\includegraphics[width=0.95\columnwidth]{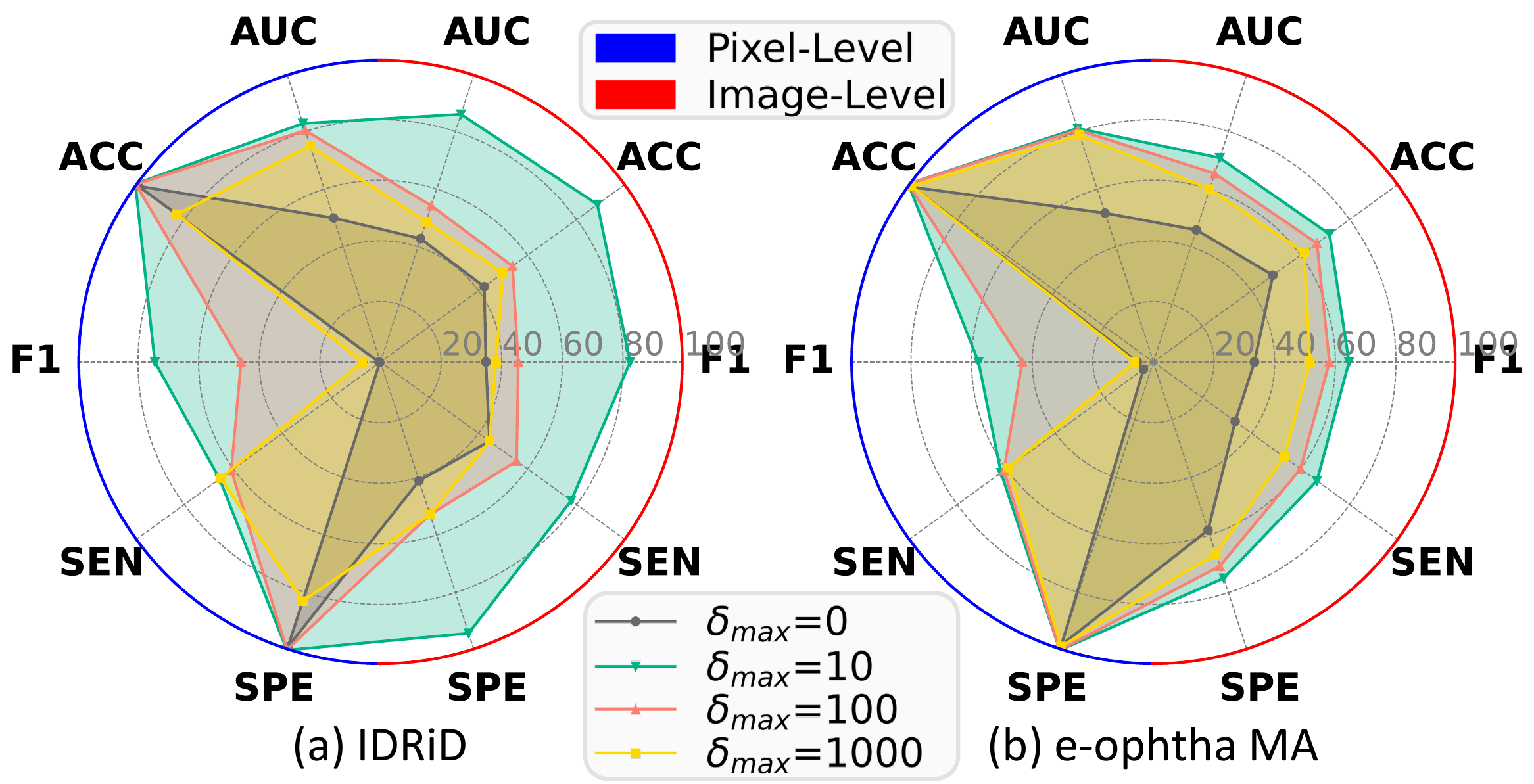}
\caption{The quantitative results for WDT-MD under different maximum noise encoding timesteps $\delta_{max}$. (Unit: \%)}
\label{delta}
\end{figure}

Furthermore, we explored the impact of varying maximum noise encoding timesteps. As depicted in Fig. \ref{delta}, our model performs best at $\delta_{max}=10$. This indicates that moderate noise encoding effectively mitigates ``identity mapping", while too large $\delta_{max}$ implies excessive noise injection and more difficult model convergence, causing performance degradation.

% In particular, pixel-level SEN is improved by 65.61\% and 62.12\% on IDRiD and e-ophtha MA, respectively. Whereas, too large $\delta_{max}$ implies excessive noise injection and more difficult model convergence, causing performance degradation. 

\section{Conclusion}
% In this paper, we propose a novel AD framework for MA detection, namely WDT-MD. To avoid ``identity mapping", we introduces a noise-encoded image conditioning mechanism. Considering the high false positive dilemma, we employ pseudo-normal pattern synthesis to introduce pixel-level supervision, enabling discrimination between MAs and other anomalies. Furthermore, by combining the global modeling capability of DiTs with multi-scale wavelet analysis, our wavelet diffusion Transformer architecture enhances the reconstruction quality of normal retinal features. Extensive experiments demonstrate that WDT-MD outperforms other methods in MA detection. In future work, we will explore integration with multimodal ophthalmic data to further expedite clinical adoption of AI-powered early DR screening.

In this paper, we introduce WDT-MD, a novel supervised AD framework for MA detection. WDT-MD incorporates the noise-encoded image conditioning mechanism to mitigate ``identity mapping", pseudo-normal pattern synthesis to reduce false positives, and the wavelet diffusion Transformer architecture to enhance reconstruction quality of normal retinal features. Extensive experiments demonstrate WDT-MD's superior performance. In future work, we will explore integration with multimodal ophthalmic data to further expedite clinical adoption of AI-powered early DR screening.

\section{Acknowledgments}
This work was supported by the National Key Research and Development Program of China (No. 2024ZD0536605, No. 2025YFE0103200), Transvascular Implantation Devices Research Institute (TIDRI) (No. KY052025008), National Natural Science Foundation of China (No. 61702146, No. 62076084, No. U20A20386, No. U22A2033, No. 62302399), Zhejiang Key Research and Development Program of China (No. 2024SSYS0026), Zhejiang Key Laboratory of Medical Imaging Artificial Intelligence, Zhejiang Provincial Natural Science Foundation of China (No. LY21F020017, No. 2023C03090), GuangDong Basic and Applied Basic Research Foundation (No. 2025A1515011617), and National Undergraduate Innovation and Entrepreneurship Training Program of China (No. 202510336076, No. 202410336081).

\bibliography{main}

\begin{thebibliography}{46}
\providecommand{\natexlab}[1]{#1}

\bibitem[{Arrigo et~al.(2024)Arrigo, Aragona, Teussink, Battaglia~Parodi, and Bandello}]{arrigo2024digital}
Arrigo, A.; Aragona, E.; Teussink, M.; Battaglia~Parodi, M.; and Bandello, F. 2024.
\newblock Digital histology of retinal microaneurysms as provided by dense B-scan (DART) OCTA: Characteristics and clinical relevance in diabetic retinopathy.
\newblock \emph{Eye}, 38(16): 3108--3112.

\bibitem[{Baitieva et~al.(2024)Baitieva, Hurych, Besnier, and Bernard}]{baitieva2024supervised}
Baitieva, A.; Hurych, D.; Besnier, V.; and Bernard, O. 2024.
\newblock Supervised anomaly detection for complex industrial images.
\newblock In \emph{Proceedings of the IEEE/CVF Conference on Computer Vision and Pattern Recognition}, 17754--17762.

\bibitem[{Bao et~al.(2023)Bao, Nie, Xue, Cao, Li, Su, and Zhu}]{bao2023all}
Bao, F.; Nie, S.; Xue, K.; Cao, Y.; Li, C.; Su, H.; and Zhu, J. 2023.
\newblock All are worth words: A vit backbone for diffusion models.
\newblock In \emph{Proceedings of the IEEE/CVF Conference on Computer Vision and Pattern Recognition}, 22669--22679.

\bibitem[{Baugh et~al.(2024)Baugh, Reynaud, Marimont, Cechnicka, M{\"u}ller, Tarroni, and Kainz}]{baugh2024image}
Baugh, M.; Reynaud, H.; Marimont, S.~N.; Cechnicka, S.; M{\"u}ller, J.~P.; Tarroni, G.; and Kainz, B. 2024.
\newblock Image-conditioned diffusion models for medical anomaly detection.
\newblock In \emph{International Workshop on Uncertainty for Safe Utilization of Machine Learning in Medical Imaging}, 117--127. Springer.

\bibitem[{Cai, Chen, and Cheng(2024)}]{cai2024rethinking}
Cai, Y.; Chen, H.; and Cheng, K.-T. 2024.
\newblock Rethinking autoencoders for medical anomaly detection from a theoretical perspective.
\newblock In \emph{International Conference on Medical Image Computing and Computer-Assisted Intervention}, 544--554. Springer.

\bibitem[{Decenciere et~al.(2013)Decenciere, Cazuguel, Zhang, Thibault, Klein, Meyer, Marcotegui, Quellec, Lamard, Danno et~al.}]{decenciere2013teleophta}
Decenciere, E.; Cazuguel, G.; Zhang, X.; Thibault, G.; Klein, J.-C.; Meyer, F.; Marcotegui, B.; Quellec, G.; Lamard, M.; Danno, R.; et~al. 2013.
\newblock TeleOphta: Machine learning and image processing methods for teleophthalmology.
\newblock \emph{IRBM}, 34(2): 196--203.

\bibitem[{Dhariwal and Nichol(2021)}]{dhariwal2021diffusion}
Dhariwal, P.; and Nichol, A. 2021.
\newblock Diffusion models beat gans on image synthesis.
\newblock \emph{Advances in Neural Information Processing Systems}, 34: 8780--8794.

\bibitem[{Esser et~al.(2024)Esser, Kulal, Blattmann, Entezari, M{\"u}ller, Saini, Levi, Lorenz, Sauer, Boesel et~al.}]{esser2024scaling}
Esser, P.; Kulal, S.; Blattmann, A.; Entezari, R.; M{\"u}ller, J.; Saini, H.; Levi, Y.; Lorenz, D.; Sauer, A.; Boesel, F.; et~al. 2024.
\newblock Scaling rectified flow transformers for high-resolution image synthesis.
\newblock In \emph{International Conference on Machine Learning}, 12606--12633. PMLR.

\bibitem[{Feng et~al.(2025)Feng, Ma, Wang, Qi, Chen, Chen, and Wang}]{feng2025dit4edit}
Feng, K.; Ma, Y.; Wang, B.; Qi, C.; Chen, H.; Chen, Q.; and Wang, Z. 2025.
\newblock Dit4edit: Diffusion transformer for image editing.
\newblock In \emph{Proceedings of the AAAI Conference on Artificial Intelligence}, volume~39, 2969--2977.

\bibitem[{Fontanella et~al.(2024)Fontanella, Mair, Wardlaw, Trucco, and Storkey}]{fontanella2024diffusion}
Fontanella, A.; Mair, G.; Wardlaw, J.; Trucco, E.; and Storkey, A. 2024.
\newblock Diffusion models for counterfactual generation and anomaly detection in brain images.
\newblock \emph{IEEE Transactions on Medical Imaging}.

\bibitem[{Foo, Hsu, and Lee(2023)}]{foo2023multi}
Foo, A.; Hsu, W.; and Lee, M.~L. 2023.
\newblock Multi-object representation learning via feature connectivity and object-centric regularization.
\newblock \emph{Advances in Neural Information Processing Systems}, 36: 60035--60047.

\bibitem[{Goodfellow et~al.(2014)Goodfellow, Pouget-Abadie, Mirza, Xu, Warde-Farley, Ozair, Courville, and Bengio}]{goodfellow2014generative}
Goodfellow, I.~J.; Pouget-Abadie, J.; Mirza, M.; Xu, B.; Warde-Farley, D.; Ozair, S.; Courville, A.; and Bengio, Y. 2014.
\newblock Generative adversarial nets.
\newblock \emph{Advances in Neural Information Processing Systems}, 27.

\bibitem[{Guo et~al.(2025)Guo, Lu, Zhang, Chen, Li, and Liao}]{guo2025dinomaly}
Guo, J.; Lu, S.; Zhang, W.; Chen, F.; Li, H.; and Liao, H. 2025.
\newblock Dinomaly: The less is more philosophy in multi-class unsupervised anomaly detection.
\newblock In \emph{Proceedings of the Computer Vision and Pattern Recognition Conference}, 20405--20415.

\bibitem[{Ho, Jain, and Abbeel(2020)}]{ho2020denoising}
Ho, J.; Jain, A.; and Abbeel, P. 2020.
\newblock Denoising diffusion probabilistic models.
\newblock \emph{Advances in Neural Information Processing Systems}, 33: 6840--6851.

\bibitem[{Huang et~al.(2024)Huang, Huang, Liu, Yan, Dong, Lv, Chen, and Chen}]{huang2024wavedm}
Huang, Y.; Huang, J.; Liu, J.; Yan, M.; Dong, Y.; Lv, J.; Chen, C.; and Chen, S. 2024.
\newblock Wavedm: Wavelet-based diffusion models for image restoration.
\newblock \emph{IEEE Transactions on Multimedia}, 26: 7058--7073.

\bibitem[{Jiang et~al.(2024)Jiang, Gao, Liu, Tang, Zhang, Jiang, Yuan, and Liu}]{jiang2024glanceseg}
Jiang, H.; Gao, M.; Liu, Z.; Tang, C.; Zhang, X.; Jiang, S.; Yuan, W.; and Liu, J. 2024.
\newblock GlanceSeg: Real-time microaneurysm lesion segmentation with gaze-map-guided foundation model for early detection of diabetic retinopathy.
\newblock \emph{IEEE Journal of Biomedical and Health Informatics}.

\bibitem[{Jiang et~al.(2023)Jiang, Hou, Miao, Ye, Gao, Li, Jin, and Liu}]{jiang2023eye}
Jiang, H.; Hou, Y.; Miao, H.; Ye, H.; Gao, M.; Li, X.; Jin, R.; and Liu, J. 2023.
\newblock Eye tracking based deep learning analysis for the early detection of diabetic retinopathy: A pilot study.
\newblock \emph{Biomedical Signal Processing and Control}, 84: 104830.

\bibitem[{Kascenas et~al.(2023)Kascenas, Sanchez, Schrempf, Wang, Clackett, Mikhael, Voisey, Goatman, Weir, Pugeault et~al.}]{kascenas2023role}
Kascenas, A.; Sanchez, P.; Schrempf, P.; Wang, C.; Clackett, W.; Mikhael, S.~S.; Voisey, J.~P.; Goatman, K.; Weir, A.; Pugeault, N.; et~al. 2023.
\newblock The role of noise in denoising models for anomaly detection in medical images.
\newblock \emph{Medical Image Analysis}, 90: 102963.

\bibitem[{Khan et~al.(2025)Khan, Gaidhane, Singh, Ganesan, Kaur, Sharma, Rani, Sharma, Thapliyal, Kushwaha et~al.}]{khan2025diagnostic}
Khan, Z.; Gaidhane, A.~M.; Singh, M.; Ganesan, S.; Kaur, M.; Sharma, G.~C.; Rani, P.; Sharma, R.; Thapliyal, S.; Kushwaha, M.; et~al. 2025.
\newblock Diagnostic accuracy of IDX-DR for detecting diabetic retinopathy: A systematic review and Meta-Analysis.
\newblock \emph{American Journal of Ophthalmology}, 273: 192--204.

\bibitem[{Kumar et~al.(2025)Kumar, Chakraborty, Mahapatra, Bozorgtabar, and Roy}]{kumar2025self}
Kumar, K.; Chakraborty, S.; Mahapatra, D.; Bozorgtabar, B.; and Roy, S. 2025.
\newblock Self-supervised anomaly segmentation via diffusion models with dynamic Transformer UNet.
\newblock In \emph{Proceedings of the IEEE/CVF Winter Conference on Applications of Computer Vision}, 7928--7938.

\bibitem[{Li et~al.(2024)Li, Feng, Chen, Chen, Wang, Hu, Sun, Qu, and Zhou}]{li2024vague}
Li, Y.; Feng, Y.; Chen, B.; Chen, W.; Wang, Y.; Hu, X.; Sun, B.; Qu, C.; and Zhou, M. 2024.
\newblock Vague prototype-oriented diffusion model for multi-class anomaly detection.
\newblock In \emph{International Conference on Machine Learning}, 27771--27790. PMLR.

\bibitem[{Luo et~al.(2023)Luo, Tan, Huang, Li, and Zhao}]{luo2023latent}
Luo, S.; Tan, Y.; Huang, L.; Li, J.; and Zhao, H. 2023.
\newblock Latent consistency models: Synthesizing high-resolution images with few-step inference.
\newblock \emph{arXiv preprint arXiv:2310.04378}.

\bibitem[{Ma et~al.(2024{\natexlab{a}})Ma, Goldstein, Albergo, Boffi, Vanden-Eijnden, and Xie}]{ma2024sit}
Ma, N.; Goldstein, M.; Albergo, M.~S.; Boffi, N.~M.; Vanden-Eijnden, E.; and Xie, S. 2024{\natexlab{a}}.
\newblock Sit: Exploring flow and diffusion-based generative models with scalable interpolant transformers.
\newblock In \emph{European Conference on Computer Vision}, 23--40. Springer.

\bibitem[{Ma et~al.(2025{\natexlab{a}})Ma, He, Wang, Wang, Shen, Qi, Ying, Cai, Li, Shum et~al.}]{ma2025followyourclick}
Ma, Y.; He, Y.; Wang, H.; Wang, A.; Shen, L.; Qi, C.; Ying, J.; Cai, C.; Li, Z.; Shum, H.-Y.; et~al. 2025{\natexlab{a}}.
\newblock Follow-your-click: Open-domain regional image animation via motion prompts.
\newblock In \emph{Proceedings of the AAAI Conference on Artificial Intelligence}, volume~39, 6018--6026.

\bibitem[{Ma et~al.(2024{\natexlab{b}})Ma, Liu, Wang, Pan, He, Yuan, Zeng, Cai, Shum, Liu et~al.}]{ma2024followyouremoji}
Ma, Y.; Liu, H.; Wang, H.; Pan, H.; He, Y.; Yuan, J.; Zeng, A.; Cai, C.; Shum, H.-Y.; Liu, W.; et~al. 2024{\natexlab{b}}.
\newblock Follow-your-emoji: Fine-controllable and expressive freestyle portrait animation.
\newblock In \emph{SIGGRAPH Asia 2024 Conference Papers}, 1--12.

\bibitem[{Ma et~al.(2025{\natexlab{b}})Ma, Yan, Liu, Wang, Pan, He, Yuan, Zeng, Cai, Shum et~al.}]{ma2025followfaster}
Ma, Y.; Yan, Z.; Liu, H.; Wang, H.; Pan, H.; He, Y.; Yuan, J.; Zeng, A.; Cai, C.; Shum, H.-Y.; et~al. 2025{\natexlab{b}}.
\newblock Follow-your-emoji-faster: Towards efficient, fine-controllable, and expressive freestyle portrait animation.
\newblock \emph{arXiv preprint arXiv:2509.16630}.

\bibitem[{Mayya, Kamath, and Kulkarni(2021)}]{mayya2021automated}
Mayya, V.; Kamath, S.; and Kulkarni, U. 2021.
\newblock Automated microaneurysms detection for early diagnosis of diabetic retinopathy: A comprehensive review.
\newblock \emph{Computer Methods and Programs in Biomedicine Update}, 1: 100013.

\bibitem[{Peebles and Xie(2023)}]{peebles2023scalable}
Peebles, W.; and Xie, S. 2023.
\newblock Scalable diffusion models with transformers.
\newblock In \emph{Proceedings of the IEEE/CVF International Conference on Computer Vision}, 4195--4205.

\bibitem[{Porwal et~al.(2018)Porwal, Pachade, Kamble, Kokare, Deshmukh, Sahasrabuddhe, and Meriaudeau}]{porwal2018indian}
Porwal, P.; Pachade, S.; Kamble, R.; Kokare, M.; Deshmukh, G.; Sahasrabuddhe, V.; and Meriaudeau, F. 2018.
\newblock Indian diabetic retinopathy image dataset (IDRiD): a database for diabetic retinopathy screening research.
\newblock \emph{Data}, 3(3): 25.

\bibitem[{Porwal et~al.(2020)Porwal, Pachade, Kokare, Deshmukh, Son, Bae, Liu, Wang, Liu, Gao et~al.}]{porwal2020idrid}
Porwal, P.; Pachade, S.; Kokare, M.; Deshmukh, G.; Son, J.; Bae, W.; Liu, L.; Wang, J.; Liu, X.; Gao, L.; et~al. 2020.
\newblock Idrid: Diabetic retinopathy--segmentation and grading challenge.
\newblock \emph{Medical Image Analysis}, 59: 101561.

\bibitem[{Raghu et~al.(2019)Raghu, Zhang, Kleinberg, and Bengio}]{raghu2019transfusion}
Raghu, M.; Zhang, C.; Kleinberg, J.; and Bengio, S. 2019.
\newblock Transfusion: Understanding transfer learning for medical imaging.
\newblock \emph{Advances in Neural Information Processing Systems}, 32.

\bibitem[{Shao et~al.(2025{\natexlab{a}})Shao, Miao, Duan, Wang, Chen, Huang, Wu, Deng, Long, and Zheng}]{shao2025trace}
Shao, M.; Miao, X.; Duan, H.; Wang, Z.; Chen, J.; Huang, Y.; Wu, X.; Deng, J.; Long, Y.; and Zheng, Y. 2025{\natexlab{a}}.
\newblock Trace: Temporally reliable anatomically-conditioned 3D CT generation with enhanced efficiency.
\newblock In \emph{International Conference on Medical Image Computing and Computer-Assisted Intervention}, 627--637. Springer.

\bibitem[{Shao et~al.(2025{\natexlab{b}})Shao, Wang, Duan, Huang, Zhai, Wang, Long, and Zheng}]{shao2025rethinking}
Shao, M.; Wang, Z.; Duan, H.; Huang, Y.; Zhai, B.; Wang, S.; Long, Y.; and Zheng, Y. 2025{\natexlab{b}}.
\newblock Rethinking brain tumor segmentation from the frequency domain perspective.
\newblock \emph{IEEE Transactions on Medical Imaging}.

\bibitem[{Sun et~al.(2025{\natexlab{a}})Sun, Cao, Dong, and Fink}]{sun2025unseen}
Sun, H.; Cao, Y.; Dong, H.; and Fink, O. 2025{\natexlab{a}}.
\newblock Unseen visual anomaly generation.
\newblock In \emph{Proceedings of the Computer Vision and Pattern Recognition Conference}, 25508--25517.

\bibitem[{Sun et~al.(2025{\natexlab{b}})Sun, Chen, Zheng, Ge, Liu, Min, Elazab, Wan, and Wang}]{sun2025bs}
Sun, Y.; Chen, Z.; Zheng, H.; Ge, R.; Liu, J.; Min, W.; Elazab, A.; Wan, X.; and Wang, C. 2025{\natexlab{b}}.
\newblock Bs-ldm: Effective bone suppression in high-resolution chest X-ray images with conditional latent diffusion models.
\newblock \emph{IEEE Journal of Biomedical and Health Informatics}.

\bibitem[{Sun et~al.(2025{\natexlab{c}})Sun, Chen, Zheng, Lu, Duan, Fan, Elazab, Wan, Wang, and Ge}]{sun2025gl}
Sun, Y.; Chen, Z.; Zheng, H.; Lu, Y.; Duan, L.; Fan, F.; Elazab, A.; Wan, X.; Wang, C.; and Ge, R. 2025{\natexlab{c}}.
\newblock Gl-lcm: Global-local latent consistency models for fast high-resolution bone suppression in chest X-ray images.
\newblock In \emph{International Conference on Medical Image Computing and Computer-Assisted Intervention}, 222--232. Springer.

\bibitem[{Telea(2004)}]{telea2004image}
Telea, A. 2004.
\newblock An image inpainting technique based on the fast marching method.
\newblock \emph{Journal of Graphics Tools}, 9(1): 23--34.

\bibitem[{Wang et~al.(2024)Wang, Bai, Li, Zhai, Jiang, and Liu}]{wang2024learning}
Wang, K.; Bai, Y.; Li, D.; Zhai, D.; Jiang, J.; and Liu, X. 2024.
\newblock Learning lossless compression for high bit-depth volumetric medical image.
\newblock \emph{IEEE Transactions on Image Processing}.

\bibitem[{Wang, Xu, and Zhao(2024)}]{wang2024wavelet}
Wang, Y.; Xu, J.; and Zhao, Y. 2024.
\newblock Wavelet encoding network for inertial signal enhancement via feature supervision.
\newblock \emph{IEEE Transactions on Industrial Informatics}.

\bibitem[{Wolleb et~al.(2022)Wolleb, Bieder, Sandk{\"u}hler, and Cattin}]{wolleb2022diffusion}
Wolleb, J.; Bieder, F.; Sandk{\"u}hler, R.; and Cattin, P.~C. 2022.
\newblock Diffusion models for medical anomaly detection.
\newblock In \emph{International Conference on Medical Image Computing and Computer-Assisted Intervention}, 35--45. Springer.

\bibitem[{Wu and Jiao(2024)}]{wu2024microseg}
Wu, Y.; and Jiao, G. 2024.
\newblock MicroSeg: Multi-scale fusion learning for microaneurysms segmentation.
\newblock \emph{Biomedical Signal Processing and Control}, 97: 106700.

\bibitem[{Wyatt et~al.(2022)Wyatt, Leach, Schmon, and Willcocks}]{wyatt2022anoddpm}
Wyatt, J.; Leach, A.; Schmon, S.~M.; and Willcocks, C.~G. 2022.
\newblock Anoddpm: Anomaly detection with denoising diffusion probabilistic models using simplex noise.
\newblock In \emph{Proceedings of the IEEE/CVF Conference on Computer Vision and Pattern Recognition}, 650--656.

\bibitem[{Xu et~al.(2024)Xu, Luo, Huang, Liu, Wen, Wang, and Xu}]{xu2024hacdr}
Xu, Q.; Luo, X.; Huang, C.; Liu, C.; Wen, J.; Wang, J.; and Xu, Y. 2024.
\newblock HACDR-Net: heterogeneous-aware convolutional network for diabetic retinopathy multi-lesion segmentation.
\newblock In \emph{Proceedings of the AAAI Conference on Artificial Intelligence}, 6342--6350.

\bibitem[{Yap and Ng(2023)}]{yap2023cut}
Yap, B.~P.; and Ng, B.~K. 2023.
\newblock Cut-paste consistency learning for semi-supervised lesion segmentation.
\newblock In \emph{Proceedings of the IEEE/CVF Winter Conference on Applications of Computer Vision}, 6160--6169.

\bibitem[{Zhang et~al.(2024)Zhang, Liu, Xie, Huang, Zhang, Li, Ramachandra, and Zheng}]{zhang2024anomaly}
Zhang, W.; Liu, H.; Xie, J.; Huang, Y.; Zhang, Y.; Li, Y.; Ramachandra, R.; and Zheng, Y. 2024.
\newblock Anomaly detection via gating highway connection for retinal fundus images.
\newblock \emph{Pattern Recognition}, 148: 110167.

\bibitem[{Zhao et~al.(2024)Zhao, Cai, Dong, and Hu}]{zhao2024wavelet}
Zhao, C.; Cai, W.; Dong, C.; and Hu, C. 2024.
\newblock Wavelet-based fourier information interaction with frequency diffusion adjustment for underwater image restoration.
\newblock In \emph{Proceedings of the IEEE/CVF Conference on Computer Vision and Pattern Recognition}, 8281--8291.

\end{thebibliography}

\end{document}